\documentclass[sigconf]{acmart}

\usepackage{algorithm}
\usepackage{algpseudocode}
\usepackage{array}
\usepackage{amsmath, amssymb, enumitem}
\usepackage{fixltx2e}
\usepackage{url}
\usepackage{hyperref}
\usepackage{units}
\usepackage{tikz}

\usepackage{xcolor}
\usepackage{hhline}
\usepackage{gensymb}
\usepackage{currvita}
\usepackage{xfrac}
\usepackage{subfig}
\usepackage{graphicx}
\usepackage{lipsum}

\usepackage{textcomp}

\newcommand{\bmm}[1]{\ensuremath{\mathbf{#1}}}

\newcommand*\circled[1]{\tikz[baseline=(char.base)]{
            \node[shape=circle,fill,inner sep=1pt] (char) {\textcolor{white}{#1}};}}

\copyrightyear{2020}
\acmYear{2020}
\setcopyright{none}
\acmConference[ISLPED '20]

\begin{document}

\title{SHEAR$er$: Highly-Efficient Hyperdimensional Computing by \underline{S}oftware-\underline{H}ardware \underline{E}nabled Multifold \underline{A}pp\underline{R}oximation}

\author{Behnam Khaleghi, Sahand Salamat, Anthony Thomas, Fatemeh Asgarinejad, Yeseong Kim, and Tajana Rosing}
\affiliation{Computer Science and Engineering Department, UC San Diego, La Jolla, CA 92093, USA\\
    \{bkhaleghi, sasalama, ahthomas, fasgarinejad, yek048, tajana\}@ucsd.edu }

\begin{abstract}
Hyperdimensional computing (HD) is an emerging paradigm for machine learning based on the evidence that the brain computes on high-dimensional, distributed, representations of data.
The main operation of HD is encoding, which transfers the input data to hyperspace by mapping each input feature to a hypervector, accompanied by so-called bundling procedure that simply adds up the hypervectors to realize encoding hypervector.
Although the operations of HD are highly parallelizable, the massive number of operations hampers the efficiency of HD in embedded domain.
In this paper, we propose SHEAR$er$, an algorithm-hardware co-optimization to improve the performance and energy consumption of HD computing.
We gain insight from a prudent scheme of approximating the hypervectors that, thanks to inherent error resiliency of HD, has minimal impact on accuracy while provides high prospect for hardware optimization.
In contrast to previous works that generate the encoding hypervectors in full precision and then \emph{ex-post} quantizing, we compute the encoding hypervectors in an approximate manner that saves a significant amount of resources yet affords high accuracy.
We also propose a novel FPGA implementation that achieves striking performance through massive parallelism with low power consumption.
Moreover, we develop a software framework that enables training HD models by emulating the proposed approximate encodings.
The FPGA implementation of SHEAR$er$ achieves an average throughput boost of 104,904$\times$ (15.7$\times$) and energy savings of up to 56,044$\times$ (301$\times$) compared to state-of-the-art encoding methods implemented on Raspberry Pi 3 (GeForce GTX 1080 Ti) using practical machine learning datasets.
\end{abstract}

\maketitle


\section{Introduction} \label{sec:intro}
Networked sensors with native computing power -- otherwise known as the ``internet of things'' (IoT) -- are a rapidly growing source of data. Applications based on IoT devices typically use machine learning (ML) algorithms to generate useful insights from data. While modern machine learning techniques -- in particular deep neural networks (DNNs) -- can produce state-of-the-art results, they often entail substantial memory and compute requirements which may exceed the resources available on light-weight edge devices. Thus, there is a pressing need to develop novel machine learning techniques which provide accuracy and flexibility while meeting the tight resource constraints imposed by edge-sensing devices.  

Hyperdimensional computing -- HD for short -- is an emerging paradigm for machine learning based on evidence from the neuroscience community that the brain ``computes'' on high-dimensional, distributed, representations of data \cite{kanerva2009hyperdimensional,masse2009olfactory,turner2008olfactory,wilson2013early,olshausen2004sparse}. In HD, the primitive units of computation are high-dimensional vectors of length $d_{hv}$ sampled randomly from the uniform distribution over the binary cube $\{\pm 1\}^{d_{hv}}$. Typical values of $d_{hv}$ are in the range 5-10,000. Because of their high-dimensionality, any randomly chosen pair of points will be approximately orthogonal (that is, their inner product will be approximately zero). A useful consequence of this is that sets can be encoded simply by summing (or ``bundling'') together their constituent vectors. For any collection of vectors $\bmm{P},\bmm{Q},\bmm{V}$ their element-wise sum $\bmm{S}=\bmm{P}+\bmm{Q}+\bmm{V}$ is, in expectation, closer to $\bmm{P}, \bmm{Q}$ and $\bmm{V}$ than any other randomly chosen vector in the space.

\begin{figure}[t]
  \centering
  \includegraphics[width=1.0\columnwidth]{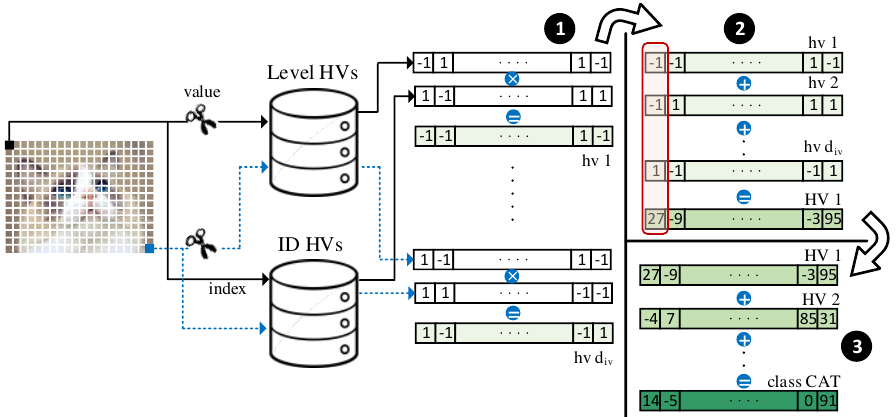} \vspace{-0.0cm}
  \caption{Encoding and training in HD.}\label{fig:HD}  \vspace{-0.0cm}
\end{figure}

Given HD representations of data, this provides a simple classification scheme: we simply take the data points corresponding to a particular class and superimpose them into a single representation for the set. Then, given a new piece of data for which the correct class label is unknown, we compute the similarity with the hypervectors representing each class and return the label corresponding to the most similar one. More formally, suppose we are given a set of labeled data $\mathcal{X} = \{(\bmm{x}_{i}, y_{i})\}_{i=1}^{N}$ where $\bmm{x} \in \mathbb{R}^{d_{iv}}$ corresponds to an observation in low-dimensional space and $y \in \mathcal{C}$ is a categorical variable indicating the class to which a particular $\bmm{x}$ belongs. In general, HD classification proceeds by generating a set of ``class hypervectors'' which represent the training data corresponding to each class. Then, given a piece of data for which we do not know the correct label -- the ``query'' -- we simply compute the similarity between the query and each class hypervector and return the label corresponding to the most similar. This process is illustrated in Figure \ref{fig:HD}.

Suppose we wish to generate the class hypervector corresponding to some class $k \in \mathcal{C}$. The prototype can be generated simply by superimposing (also called ``bundling'' in the literature) the HD-encoded representation of the training data corresponding to that particular class \cite{plate1995holographic,kanerva2009hyperdimensional}:
\begin{gather}
    \label{eqn:classification}
    \bmm{C}_{k} = \sum_{i \text{ s.t. } y_{i} = k} \text{enc}(\bmm{x}_{i})
\end{gather}
where $\text{enc} : \mathbb{R}^{d_{iv}} \rightarrow \{\pm 1\}^{d_{hv}}$ is some \emph{encoding function} which maps a low-dimensional signal to a binary HD representation. Then, given some piece of ``query'' data $\bmm{x}_{q}$ for which we \emph{do not} know the correct label we simple return the predicted label as:
\begin{gather}
    k^{\star} = \underset{k \in \mathcal{C}}{\text{argmax }} \delta(\text{enc}(\bmm{x}_{q}),\bmm{C}_{k})
\end{gather}
where $\delta$ is an appropriate similarity metric. Common choices for $\delta$ include the inner-product/cosine distance -- appropriate for integer or real valued encoding schemes -- and the hamming distance -- appropriate for binary HD representations. This phase is commonly referred to in literature as ``associative search''.
Despite the simplicity of this ``learning'' scheme, HD computing has been successfully applied to a number of practical problems in the literature ranging from optimizing the performance of web-browsers \cite{wan2012web}, to DNA sequence alignment \cite{kim2020genie,imani2018hdna}, bio-signal processing \cite{rahimi2018efficient, fatemeh2020embc}, robotics \cite{mitrokhin2019learning,neubert2019introduction}, and privacy preserving federated learning \cite{imani2019framework, behnam2020hd}.

The primary appeal of HD computing lies in its amenability to implementation in modern hardware accelerators. Because the HD representations (e.g. $\phi(\bmm{x})$) are simply long Boolean vectors, they can be processed extremely efficiently in highly parallel platforms like GPUs, FPGAs and PIM architectures. The principal challenge of HD computing -- and the focus of this paper -- lies in designing good encoding schemes which (1) represent the data in a format suitable for learning and (2) are efficient to implement in hardware. In general, the encoding phase is the most expensive stage in the HD learning pipeline -- in some cases taking up to $10\times$ longer than training or prediction \cite{imani2019bric}. Existing encoding methods require generating hypervectors in full integer-precision and then \emph{ex-post} quantizing to $\{\pm 1\}$. While this accelerates the associative search phase, it does not address encoding which is the primary source of inefficiency. 

In this work, we propose novel techniques to compute the encodings in an approximate manner that saves a substantial amount of resources with an insignificant impact on accuracy.
Of independent interest is our novel FPGA implementation that achieves striking performance through massive parallelism with low power consumption.
Approximate encodings entail models to be trained in a similar approximate fashion. Thus we also develop a software emulation to enable users to train desired HD models.
Our software framework enables users to explore the tradeoff between the degree of approximation, accuracy, and resource utilization (hence power consumption) by generating a pre-compiled library that correlates approximation schemes and FPGA resource utilization and power consumption.
We show our procedure leads to performance improvement of 104,904$\times$ (15.7$\times$) and energy savings of up to 56,044$\times$ (301$\times$) compared to state-of-the-art encoding methods implemented on Raspberry Pi 3 (GeForce GTX 1080 Ti).

\section{Background and Motivation} \label{sec:background}

\subsection{HD Encoding Algorithms} \label{subsec:encoding}

The literature has proposed a number of encoding methods for the multitude of data types which arise in practical learning settings. We here focus on a method from \cite{plate1995holographic,kanerva2009hyperdimensional,rahimi2016robust} which we refer to as ``ID-vector'' based encoding. This encoding method is widely used (see for instance: \cite{imani2017voicehd,rahimi2016robust,imani2019sparsehd,rahimi2018efficient}) and works well on both discrete and continuous data. We focus the discussion on continuous data as discrete data is a simple extension. 

Suppose we wish to encode some set of vectors $\mathcal{X} = \{\bmm{x}_{i}\}_{i=1}^{N}$ where $\bmm{x}_{i}$ is supported on some compact subset of $\mathbb{R}^{d_{iv}}$. To begin, we first quantize the domain of each feature into a set of $L$ discrete values $\mathcal{L} = \{l_{i}\}_{i=1}^{L}$ and assign each $l_{i} \in \mathcal{L}$ a codeword $\bmm{L}_{i} \in \{\pm 1\}^{d_{hv}}$. To preserve the \emph{ordinal relationship} between the quantizer bins (the $l_{i}$), we wish the similarity between the codewords $\bmm{L}_{i}, \bmm{L}_{j}$ to be inversely proportional to distance between the corresponding quantization bins; e.g. $\delta(\bmm{L}_{i},\bmm{L}_{j}) \propto |l_{i}-l_{j}|^{-1}$. To enforce this property we generate the codeword $\bmm{L}_{1}$ corresponding to the minimal quantizer bin $l_{1}$ by sampling randomly from $\{\pm 1\}^{d_{hv}}$. The codeword for the second bin is generated by flipping $\frac{d_{hv}}{2 \cdot L}$ random coordinates in $L_{1}$. The codeword for the third bin is generated analogously from $L_{2}$ and so on. Thus, the codewords for the minimal and maximal bins are orthogonal and $\delta(\bmm{L}_{i},\bmm{L}_{j})$ decays as $|j-i|$ increases. This scheme is appropriate for quantizers with linearly spaced bins -- however, it can be extended to variable bin-width quantizers.

To complete the description of encoding, let $q(x_{i})$ be a function which returns the appropriate codeword $\bmm{L} \in \mathcal{L}$ for a component $x_{i} \in \bmm{x}$. Then encoding proceeds as follows:
\begin{gather}
    \label{eqn:encoding}
    \bmm{X} = \sum_{j=i}^{d_{iv}} q(x_{i}) \otimes \bmm{P}_{i}
\end{gather}
Where $\bmm{P}_{i}$ is a ``position hypervector'' which encodes the index of the feature value (e.g. $i \in \{1,..,d_{iv}\}$) and $\otimes$ is a ``binding'' operation which is typically taken to be \texttt{XOR}.

\subsection{Motivation}

While the basic operations of HD are simple, they are numerous due to its high-dimensional nature. Prior work has proposed varied algorithmic and hardware innovations to tackle the computational challenges of HD. Acceleration in hardware has typically focused on FPGAs \cite{salamat2019f5, imani2019quanthd, schmuck2019hardware} or ASIC-ish accelerators \cite{imani2019binary, datta2019programmable}. FPGA-based implementations provide a high degree of parallelism and bit-level granularity of operations that significantly improves the performance and effective utilization of resources. Furthermore, FPGAs are advantageous over more specialized ASICs as they allow for easy customization of model parameters such as lengths of hypervectors ($d_{hv}$) and input-vectors ($d_{iv}$) along with the number of quantization levels. This flexibility is important as learning applications are heterogeneous in practice. Accordingly, we here focus on an FPGA based implementation but emphasize our techniques are generic and can be integrated with ASIC- \cite{imani2019binary} and processor-based \cite{datta2019programmable} implementations.

\begin{figure}[t]
  \centering
  \vspace{-0.0cm}
  \includegraphics[width=1.0\columnwidth]{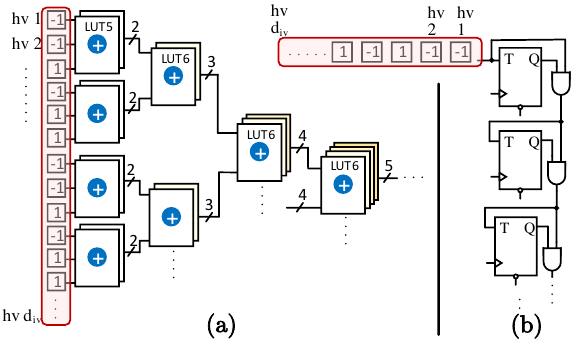} 
  \caption{(a) Adder-tree and (b) counter-based implementation of popcount. \textcircled{+} denotes add operation.}\label{fig:adder}  \vspace{-0.0cm}
\end{figure}

As noted in the preceding section, the element-wise sum is a critical operation in the encoding pipeline. Thus, popcount operations play a critical role in determining the efficiency of HD computing. Figure \ref{fig:adder}(a) shows a popular tree-based implementation of popcount that adds $d_{iv}$ binary bits (note that we can replace `$-1$'s by 0 in the hardware). Each six-input look-up table (LUT-6) of conventional FPGAs consists of two LUT-5.
Hence, we can implement the first stage of the tree using $\frac{d_{iv}}{3}$ of three-port one-bit adders.
Each subsequent stage comprises two-port $k$-bit adders where $k$ increases by one at each stage, while the number of adders per stage decreases by a factor of $\frac{1}{2}$. A $n$-bit adder requires $n$ LUT-6. Thus, the number of LUT-6 for a $d_{iv}$-input popcount can be formulated as Equation \eqref{eq:popcount}.
\begin{equation}\label{eq:popcount}
n_{\text{LUT6}}(\text{adder-tree}) = \sum_{i=1}^{\log d_{iv}}{\frac{d_{iv}}{3 } \times \frac{i}{2^{i-1}} \simeq \frac{4}{3} d_{iv}}
\end{equation}

\begin{figure*}[t]
  \centering
  \vspace{-0.0cm}
  \includegraphics[width=1.0\textwidth]{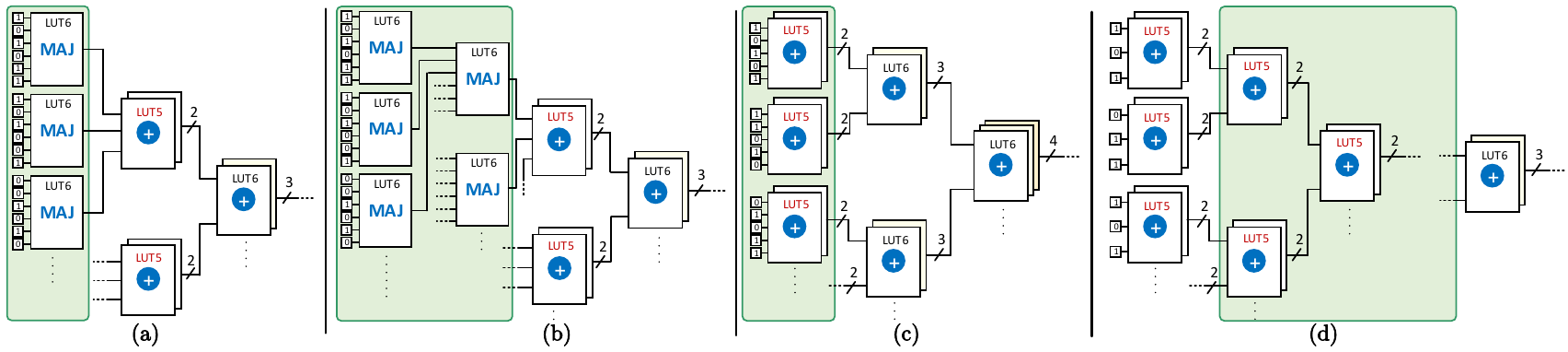} 
  \caption{Our proposed approximate encoding techniques. \texttt{MAJ} and \textcircled{+} denote majority and addition, respectively.}\label{fig:Appr}  \vspace{-0.0cm}
\end{figure*}

HD operations can be parallelized at the granularity of a single coordinate in each hypervector: all dimensions of the encoding hypervector and associative search can be computed in parallel. Nonetheless, Equation \eqref{eq:popcount} reveals that the popcount module for a popular benchmark dataset \cite{isolet} with 617 features per input requires $\sim$820 LUTs. This limits a mid-size low-power FPGA with $\sim$50K LUTs \cite{xilinx7} to generate only $\sim$60 encoding dimension per cycle (out of $d_{hv} \simeq \text{5,000}$).

To save resources, \cite{schmuck2019hardware} and \cite{imani2019binary} suggest using counters to implement the popcount for each dimension of encoding, as shown in Figure \ref{fig:adder} (b).
Although this seems more compact, in practice, it is less efficient than an adder-tree implementation: the counter-based implementation needs ``$\log d_{iv}$'' LUTs per dimension, with a per-dimension latency of $d_{iv}$ cycles, while adder-trees require $\text{O(}\frac{4}{3} d_{iv}\text{)}$ LUTs per dimension with a per-dimension throughput of one cycle, so for a given amount of resources, the conventional adder-tree is $\frac{3}{4}\log d_{iv}\times$ more performance-efficient.

Work in \cite{salamat2019f5} and \cite{imani2019quanthd} quantize the dimensions of encoding and class hypervectors which eliminates DSP modules (or large number of cascaded LUTs) that are conventionally used for the associative search stage, since, through quantization, inner product for cosine similarity will be replaced by popcount operations in case of binary quantization, or lower-bit multiplications.
The resulting improvement is minor because the quantization is applied \textit{after} full-bit encoding. Furthermore, the multipliers of the associative search stage have input widths of $w_{enc}$ (from encoding dimensions) and $w_{class}$ (from class dimensions), so each one needs $\text{O(} w_{enc} \times w_{class} \text{)}$ LUTs. Pessimistically assuming bit-widths up to $w_{enc} = w_{class} = 16$, an extreme binary quantization can eliminate 256 LUTs required for multiplication. However, the savings are again modest at best in practice: on the benchmark dataset mentioned previously, only $\frac{w_{enc} \times w_{class}}{w_{enc} \times w_{class}+\frac{4}{3} d_{iv}} \simeq 23\%$. Therefore, in this paper, we target the popcount portion that contributes to the more significant part of resources. Indeed, \emph{ex-post} quantizing of encoding hypervectors can be orthogonal to our technique for further improvement.

\section{Proposed Method: SHEAR$er$} \label{sec:prop}

\subsection{Approximate Encoding} \label{subsec:hardware}

In the previous section, we explained prior work that applies quantization after obtaining the encoding hypervector in full bit-width. As noted there, while this approach is simple it only accelerates the associative search phase and does not improve encoding - which is often the principal bottleneck. Because the HD representation of data entails substantial redundancy and information is uniformly distributed over a large number of bits, it is robust to bit-level errors: flipping 10\% of hypervectors' bits shows virtually zero accuracy drop, while 30\% bit-error impairs the accuracy by a mere 4\% \cite{imani2017exploring}.
We leverage such resilience to improve the resource utilization through approximate encoding, as shown in Figure \ref{fig:Appr}.
In the following, we discuss each technique in greater detail and estimate its resource usage.

\textbf{(1) Local majority.}
From Equation \eqref{eq:popcount} we can observe that the number of resources (in terms of LUT-6) of the exact adder-tree to see that the complexity encoding each dimension linearly depends on the number of data features, $d_{iv}$. We, therefore, aim to reduce the number of inputs to the primary adder-tree by sub-sampling using the majority function so as to shrink the tree inputs while (approximately) extracting the information contained in the input. Note that, here, `inputs' are the binary dimensions of the level hypervectors (see Figure \ref{fig:HD} \circled{2} and Figure \ref{fig:adder}). As shown in Figure \ref{fig:Appr}(a), each LUT-6 is configured to return the majority of its six input bits. When three out of six inputs are 0/1, we break the tie by designating \textit{all} LUTs that perform majority functions of a specific encoding dimension to deterministically output 0 or 1. We specify this randomly for every dimension (i.e., an entire adder-tree) but it remains fixed for a model during the training and inference.
We choose groups of six bits as a single LUT-6 can vote for up to six inputs.
Using smaller majority groups diminishes the resource saving, especially taking the majorities adds extra LUTs.
Moreover, following the Shannon decomposition, implementing a `$k+1$'-input LUT requires two $k$-input LUTs (and a two-input multiplexer). Thus, the number of LUTs for majority groups larger than six inputs grows exponentially.

There are $\frac{d_{iv}}{6}$ \texttt{MAJ} LUTs in the first stage of Figure \ref{fig:Appr}(a), hence the number of inputs for the subsequent adder-tree reduces to $\frac{d_{iv}}{6}$.
From Equation \eqref{eq:popcount} we also know that a $k$-input adder-tree requires $\frac{4}{3}k$ LUT-6. Thus, the design of Figure \ref{fig:Appr}(a) consumes:
\begin{equation}\label{fig:appr1}
\overbrace{\frac{d_{iv}}{6}}^{\text{\texttt{MAJ} LUT-6}} + \overbrace{\frac{4}{3}\frac{d_{iv}}{6}}^{\text{adder-tree}} = \frac{7}{18}d_{iv}\ \ \text{LUT-6}
\end{equation}
This uses $1 - \frac{\sfrac{7}{18}}{\sfrac{4}{3}} = 70.8\%$ less LUT resources than an exact adder-tree.

In \cite{imani2019quanthd}, the authors report an average accuracy loss of 1.6\% by \emph{post-hoc} quantizing the encodings to binary. Thus, one might think of repeating the majority functions in the subsequent stages to obtain final one-bit encoding dimensions.
Using local majority functions is efficient, but degrades the encoding quality as majority is not associative.
In particular, the \texttt{MAJ} LUTs add another layer of approximation by breaking ties.
Thus, a so-called \texttt{MAJ}-tree causes considerable accuracy loss. 
Therefore, in our cascaded-\texttt{MAJ} design in Figure \ref{fig:Appr}(b), we limit the \texttt{MAJ} stages to the first two stages. Our cascaded-\texttt{MAJ} utilizes:
\begin{equation}
\overbrace{\frac{d_{iv}}{6}}^{\text{1st stage \texttt{MAJ}s}} + \overbrace{\frac{\sfrac{d_{iv}}{6}}{6}}^{\text{2nd stage \texttt{MAJ}s}} + \overbrace{\frac{4}{3}\frac{\sfrac{d_{iv}}{6}}{6}}^{\text{adder-tree}} = \frac{25}{108}d_{iv}\ \ \text{LUT-6}
\end{equation}
which saves $1 - \frac{\sfrac{25}{108}}{\sfrac{4}{3}} = 82.6\%$ resources compared to exact encoding.
We emphasize that a cascaded all-\texttt{MAJ} popcount needs $\simeq \sum_{i=1}{\frac{1}{6^i}} = 0.2d_{iv}$ LUTs, which saves 85.0\% of LUTs. So the two-stage \texttt{MAJ} implementation with 82.6\% resource saving is nearly optimal because the first two stages of the exact tree were consuming the most resources. 

\textbf{(2) Input overfeeding.} In Figure \ref{fig:adder}(a) we can observe that each LUT-5 pair of the first stage computes $\overline{s_1s_0}$ = $hv_i + hv_{i+1} + hv_{i+2}$.
Since only three (out of five) inputs of them are used, these LUTs left underutilized.
With one more input, the output range will be [0$-$3], which requires three bits (outputs) to represent, so we cannot add more than three bits using two LUT-5s.
However, instead of using the LUT-5s to carry out regular addition, we can supply a pair of LUT-5s with five inputs to perform quantized/truncated addition.
For actual outputs (sum of five bits) of 0 or 1, the LUT-5 pair would produce 00 (zero); for 2 or 3 they produce 01 (one), and for 4 or 5 they produce 10 (two).
That is one LUT-5 computes the actual carry out of the five bits, and the other computes MSB of the sum.
To ensure that the synthesis tool infers a single LUT-6 for each pair, we can     directly instantiate LUT primitives.
As a LUT-6 comprises a LUT-5 pair (with shared inputs), the number of resources of Figure \ref{fig:Appr}(c) is:
\begin{equation}
\sum_{i=1}^{\log d_{iv}}{\frac{d_{iv}}{5} \times \frac{i}{2^{i-1}} \simeq \frac{4}{5} d_{iv}} \ \ \text{LUT-6}
\end{equation}
The first stage encompasses $\frac{d_{iv}}{5}$ LUT-6s, and each subsequent stage contains $i$-bit adders while their count decreases by $\frac{1}{2}\times$ at each stage.
Total number of LUTs is reduced by $1-\frac{\sfrac{4}{5}}{\sfrac{4}{3}} = 40\%$ (the same ratio of over-use of inputs).
The saving is smaller than the local majority approach but we expect higher accuracy due to intuitively more moderate imposed approximation.

\textbf{(3) Truncated nodes.} Out of $\frac{4}{3}d_{iv}$ LUTs used in an exact adder-tree, $d_{iv}$ (75\%) are used in the intermediate adder units.
More precisely, following $\frac{i}{2^{i}}$ ratio (see Equation \eqref{eq:popcount}), stages 1--4 of the adder contribute to 25\%, 25\%, 18.75\%, and 12.5\% of the total resources, respectively.
Note that, although the number of adder units halves at each stage, the area of each one increases linearly.
We avoid a blowup of adder sizes by truncating the least significant bit (LSB) of each adder.
As demonstrated in Figure \ref{fig:Appr}(d), the LSB of the second stage (which is supposed to have three-bit output) is discarded. Thus, instead of using two LUT-6s to compute $\overline{s_2s_1s_0} = \overline{a_1a_0} + \overline{b_1b_0}$, we can use two LUT-5s (equivalent to one LUT-6) to obtain $\overline{s_2s_1} = \overline{a_1a_0} + \overline{b_1b_0}$, where one LUT-5 computes $s_2$ and the other produces $s_1$ using four inputs $a_0$, $a_1$, $b_0$, and $b_1$.
Truncating the output of the second stage consequently decreases the output bit-width of the third stage by one bit as its inputs became two bits.
Thus, we can apply the LSB truncating to the third stage to implement it using two LUT-5s, as well.
We can apply the same procedure in all the consecutive nodes and implement them by only two LUT-5s. 
The output of the first stage is already two bits so we do not modify its original implementation.

We apply truncating to first stages particularly from the left side of Equation \ref{eq:popcount} we can perceive the first five stages that contribute to $\sim$90\% of the adder-tree resources. Otherwise, the decay in accuracy becomes too severe. 
Equation \eqref{eq:appr4} characterizes the resource usage of the adder-tree in which the first $k$ stages are implemented using 2-bit adders shown in Figure \ref{fig:Appr}(d) (including the stage one, which uses the primary exact mode).
\begin{equation} \label{eq:appr4}
\overbrace{\sum_{i=1}^{k}{\frac{d_{iv}}{3} \frac{1}{2^{i-1}}}}^{\text{the first $k$ stages}} + \overbrace{\sum_{i=k+1}^{\log d_{iv}}{\frac{d_{iv}}{3} \frac{i+1-k}{2^{i-1}}}}^{\text{subsequent stages}} \simeq \frac{d_{iv}}{3}(2+\frac{4}{2^k})\ \text{LUT-6}
\end{equation}
We can see that for $k=1$ -- i.e. when none of intermediate stages are truncated -- the equation returns $\frac{4}{3}d_{iv}$ which is equal to resources of an exact adder-tree.
Setting $k$ to 2, 3, and 4 achieves 25\%, 37.5\%, and 43.75\% resource saving, respectively.

\subsection{SHEAR$er$ Architecture} \label{subsec:arch}
Recall from Figure \ref{fig:HD}, that the HD encoding procedure needs to convert all input features to equivalent level hypervectors, bind them with the associated ID hypervector, and bundle together (e.g. sum) the resulting hypervectors to generate the final encoding.
FPGAs, however, contain limited logic resources as well as on-chip SRAM-based memory blocks (a.k.a BRAMs) to provide high performance with affordable power.
Previous work, therefore, break down this step into multiple cycles whereby at each cycle they process $d_{seg}$ dimensions \cite{salamat2019f5, imani2019sparsehd, imani2019fach}.
When processing dimensions $n \cdot d_{seg}$ to $(n+1) \cdot d_{seg}$, those architectures fetch the same dimensions of all $\mathcal{L}$ level hypervectors. Each of $d_{seg}$ adder-trees are augmented with $\mathcal{L}$-to-1 multiplexers in all of their $d_{iv}$ input ports, where the $k^\text{th} \leq d_{seg}$ adder-tree's multiplexers are connected to $k^\text{th}$ dimension of the fetched level hypervectors, and the (quantized) value of associated feature selects the right level dimension to pass.
The advantage of such architectures is that only $d_{seg} \cdot \mathcal{L}$ bits need to be fetched at each cycle.
However, it requires $d_{iv} \cdot d_{seq}$ multiplexers. For a modest $\mathcal{L} = 16$, which translates to 16-input multiplexers occupying four LUTs, the total number of LUTs used for multiplexers will be $4 \cdot d_{iv} \cdot d_{seq}$, the (exact) adder-trees occupy $d_{seq} \cdot \frac{4}{3}d_{iv}$ (in Equation \eqref{eq:popcount} we showed that a $d_{iv}$ input exact adder-tree uses $\frac{4}{3}d_{iv}$ LUTs).
This means that the augmented multiplexers occupy $3\times$ LUTs of the adder area.
In our approximate encoding, this ratio would be even larger as we trim the exact adder. Thus, multiplexer-based implementation overshadows the gain of approximating the adders as we need to preserve the copious multiplexers. 

\begin{figure}[t]
  \centering
  \includegraphics[width=1.0\columnwidth]{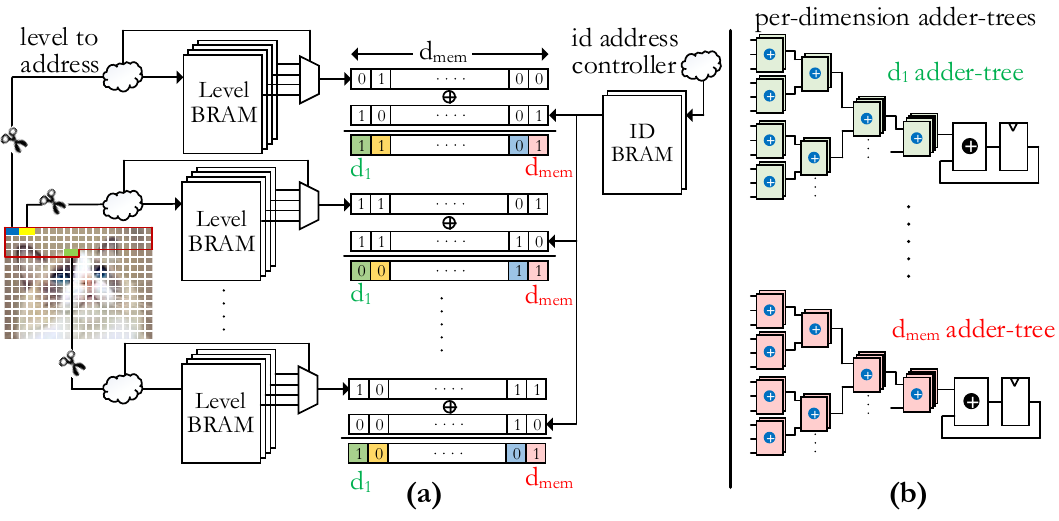} 
  \caption{SHEAR$er$ datapath abstract.}\label{fig:Arch}  \vspace{-0.0cm}
\end{figure}

To address this issue, we propose a novel FPGA implementation that relies on on-chip memories rather than adding extra resources.
Figure \ref{fig:Arch} illustrates an overview of the SHEAR$er$ FPGA architecture.
At each cycle, we partially process $\mathcal{F}$ (out of $d_{iv}$) input features, where $\mathcal{F} \leq d_{iv}$.
Our implementation is BRAM-oriented, so each (quantized) feature translates to the address from which the corresponding level hypervector can be read. This entails a dedicated memory block group for each of $\mathcal{F}$ features currently being processed.
The number of BRAMs in a group is equal to $\text{group size} = \frac{\mathcal{L} \cdot d_{hv}}{\mathcal{C}_\text{bram}}$ as there are $\mathcal{L}$ different level hypervectors of length $d_{hv}$ bits, for a memory capacity of $\mathcal{C}_\text{bram}$ bits.
Therefore, the number of features $\mathcal{F}$ that can be partially processed in a cycle is limited to $\mathcal{F} < 2 \frac{\text{total BRAMs}}{\text{group size}}$.
The coefficient 2 is because the BRAMs have two ports from which we can independently read (that is why in Figure \ref{fig:Arch} two pixels share the same BRAM group).
The address translator -- ``level to address" in Figure \ref{fig:Arch}) -- activates only the right BRAM and row of the group, so the other BRAMs do not dissipate dynamic power.
Depending on its configuration, each memory block can deliver up to $d_{mem}$ bits, as indicated in the figure.
Certainly, we could double the $d_{mem}$ by duplicating the size of memory groups to process more dimensions per cycle, but then $\mathcal{F}$ -- the number of features that can be processed -- halves.

Each of $d_{mem}$ fetched level hypervector bit is \texttt{XOR}ed with the corresponding bit of the ID (position) hypervector.
As detailed in Section \ref{subsec:encoding}, each feature \textit{index} is associated with an ID hypervector, which is a randomly chosen (but fixed) hypervector of length $d_{hv}$. We thus require $\frac{d_{hv} \cdot d_{iv}}{\mathcal{C}_\text{bram}}$ additional BRAM blocks to store ID hypervectors. This further limits the number of features that can be processed in a cycle due to BRAM shortage.
To resolve this, we only store a single ID hypervector (seed ID) and generate the other ones by rotating the seed ID, i.e., ID of index $k$ can be obtained by rotating the ID of index 1 (seed ID) by $k-1$.
This does not affect the HD accuracy as the resulting ID hypervectors are still \emph{iid} and approximately orthogonal. 
For the first feature, we need to read $d_{mem}$ bits, while for the subsequent $\mathcal{F}-1$ features we need one more bit as each ID has $d_{mem}-1$ common bits with its predecessor. Therefore we need a data-width of $d_{mem} + \mathcal{F}-1$ for ID memory, meaning that we need $1 + \frac{\mathcal{F}}{d_{mem}}$ memory blocks of the seed ID hypervector. Thus, although the seed ID fits in a single BRAM, the required data-width demands more memory blocks. However, this is still significantly smaller than the case of storing all different IDs in BRAM blocks, which either releases BRAMs for processing the features, or power gates the unused BRAMs. Moreover, using seed ID BRAM also saves dynamic power as $d_{mem} + \mathcal{F}-1$ bits are read (compared to $d_{mem} \times \mathcal{F}$ of storing different IDs).
It is also noteworthy that at each cycle the first $d_{mem}$ bits read from the ID memory are passed to the first feature of the features currently being processed (i.e., feature 1, $\mathcal{F}$ + 1, 2$\mathcal{F}$ + 1, $\cdots$). Similarly, bits 2 to $d_{mem} + 1$ of the fetched ID are passed to the second feature, and so on. Thus, the output of ID BRAMs to processing logic needs a fixed routing. 

After \texttt{XOR}ing the fetched level hypervectors with the ID hypervectors, each of the $d_{mem}$ approximate adder-trees add up $\mathcal{F}$ binary bits, so the input size of all adders is $\mathcal{F}$.
Since the result is only the sum of the first $\mathcal{F}$ features, SHEAR$er$ utilizes a buffer to store these partial sums.
In the next cycle, the procedure repeats for the next group of features, i.e., features $\mathcal{F}$ + 1 to $2 \mathcal{F}$.
Therefore, SHEAR$er$ produces $d_{mem}$ encoding dimensions in $\frac{d_{iv}}{\mathcal{F}}$ cycles, hence the entire encoding hypervector is generated in $\left \lceil{\frac{d_{hv}}{d_{mem}}}\right \rceil  \times \left \lceil{\frac{d_{iv}}{\mathcal{F}}}\right \rceil $ cycles.

To make these tangible, in the Xilinx FPGAs we use for experiments, $d_{mem}$ is 64 and $\mathcal{C}_{bram} = 512_{row} \times 64_{col}$.
We also noticed that 16 level hypervectors gives the same accuracy of having more, so we set $\mathcal{L} = 16$.
We also select the hypervector lengths to be a multiple of 512. Taking the previously mentioned language recognition benchmark \cite{isolet} as an example, we observed that $d_{hv} = \text{2,560}$ provides acceptable accuracy (see Section \ref{sec:results} for more details).
For this benchmark we thus need group size of $\left \lceil{\frac{16 \times 2560}{512 \times 64}}\right \rceil = 2$ BRAMs, where each group can cover two input features.
The FPGA we use has a total 445 BRAMs, which can make at most $\left \lfloor{\frac{445}{2}}\right \rfloor = 222$ groups, capable of processing 444 features per cycle.
Therefore, we divide 617 input features of the benchmark into two repeating cycles using 310 BRAMs (155 BRAM groups) to process the first 310 features in the first cycle, and the rest 307 cycles in the second cycle, generating $d_{mem} = 64$ encoding dimensions per 2 cycles.
All 64 adder-trees have a 1-bit input sizes of 310.
The entire encoding takes $2560\ \text{dim} \times \frac{2\ \text{cycles}}{64\ \text{dim}} = 80\ \text{cycles}$.
Note that reading from on-chip BRAMs has just one cycle latency and the off-chip memory latency is buried in the computation pipeline.


\subsection{Software Layer} \label{subsec:soft}

Because of approximation, the output of encoding and hence the class hypervectors are different than training with exact encoding.
Therefore we also need to train the model using the same approximate encoding(s), as the associative search only looks for the \textit{similarity} (rather than exactness) of an approximately encoded hypervector with trained class hypervectors -- which are made up by bundling a manifold of encoding hypervectors.
Our FPGA implementation is tailored for inference, so we carry out the training step on CPU.
We developed an efficient SIMD vectorized Python implementation to emulate the exact and the proposed encoding techniques in software.
The emulation of the proposed techniques is straightforward. For instance, for the local majority approximation (Figure \ref{fig:Appr}(a)), instead of adding up all $d_{iv}$ hypervectors, we divide them to groups of six hypervectors, add up all six hypervectors of each group, and compare if each resultant dimension is larger than 3. We also break the ties in software by generating a constant vector dictating how the ties of each dimension should be served. This acts as the \texttt{MAJ} LUTs of the first stage. Thereafter, we simply add up all these temporary hypervectors to realize the subsequent exact adders. This guarantees to match the software output with approximate hardware's, while we also achieve a fast implementation by avoiding unnecessary imitation of hardware implementation.

In addition to $d_{iv}$ that is the dataset's attribute, $d_{hv}$, $\alpha$, \texttt{epochs} (number of training epochs) are the other variables of our software implementation.
$\alpha$ is the learning rate of HD. As explained in Section \ref{sec:intro}, HD bundles all encoding hypervectors belonging to the same-label data to create the initial class hypervectors. In the subsequent \texttt{epochs} iterations, HD updates the class hypervectors by observing if the model correctly predicts the training data.
If the model mispredicts an encoded query $\mathcal{H}^l$ of label $l$ as class $\mathcal{C}^{l'}$, HD updates as shown by Equation \eqref{eq:update}. If learning rate $\alpha$ is not provided, SHEAR$er$ finds the best $\alpha$ through bisectioning for a certain number of iterations.
\begin{equation}\label{eq:update}
\begin{split}
{\mathcal{C}^l} = {\mathcal{C}^l} + \alpha \cdot \mathcal{H}^l \qquad \qquad {\mathcal{C}^{l'}} = {\mathcal{C}^{l'}} - \alpha \cdot \mathcal{H}^l
\end{split}
\end{equation}

We supply the software implementation of SHEAR$er$ with the number of BRAM and LUT resources of the target FPGA to estimate the architectural parameters according to Section \ref{subsec:arch} as well as using the resource utilization formulated in Section \ref{subsec:hardware}.
We have also implemented the exact and approximate adder-trees of different input sizes and interpolated their measured power consumption -- which is linear w.r.t. the adder size -- for different average activities of the adders' primary inputs.
Therefore, we calculate the average signal activity observed by the adders according to the values of temporary-generated binding hypervectors (level \texttt{XOR} ID). We similarly estimate the toggle rate of BRAMs according to consecutive $d_{mem}$ bits read from BRAMs.
As alluded earlier, we do not replicate the hardware implementation in software; we just need to determine each fetched level hypervector belongs to which BRAM group (based on the index of feature), so we can keep track of toggle rates.
Using the signal information with an offline look-up table created for activity-power, along with the instantiated resource information calculated as mentioned, during training, SHEAR$er$ estimates the power consumption of an application targeted for a specific device.

\section{Experimental Results} \label{sec:results}

\textbf{(1) General Setup.}
We have implemented the SHEAR$er$ architecture using Vivado High-Level Synthesis Design Suite on Xilinx Kintex-7 FPGA KC705 Evaluation Kit which embraces a XC7K325T device with 203,800 LUT-6 and 445 36 Kb BRAM memory blocks that we use in $512 \times 64$bit configuration.
By pipelining the adder-tree stages we could achieve a clock frequency of 200 MHz.
We compare the performance and energy results with the high-end NVIDIA GeForce GTX 1080 Ti GPU, and Raspberry Pi 3 embedded processor.
We optimize the CUDA implementation by packing the hypervectors within 32-bit integers, so a single logical \texttt{XOR} operation can bind 32 dimensions.
We use \texttt{speech} \cite{isolet}, \texttt{activity} \cite{ucihar}, and hand-written \texttt{digit} \cite{lecun1998mnist} recognition as well as a \texttt{face} detection dataset \cite{griffin2007caltech} as our benchmarks.
Table \ref{tab:base} summarizes the length of hypervectors and associated accuracy of each dataset in the baseline exact mode.
For a fair comparison, we first obtained the accuracies using $d_{hv} = \text{10,000}$, then decreased it until the accuracies remain within 0.5\% of the original values.
This avoids over-saturated hypervectors and accuracy drop due to approximation manifests better.

\begin{table}[t]
\caption{Baseline implementation results.} \label{tab:base}
\resizebox{0.48\textwidth}{!}{
\begin{tabular}{lcccc}
\hline
Parameter $\downarrow$\quad Benchmark $\rightarrow$  & \texttt{speech} & \texttt{activity} & \texttt{face} & \texttt{digit} \\ \hline
Input features ($d_{iv}$)                   & 617      & 561        & 608    & 784     \\
Hypervector length ($d_{hv}$)                   & 2,560      & 3,072        & 6,144    & 2,048     \\
Baseline accuracy                & 93.18\%      & 93.91\%       & 95.47\%   & 89.07\%    \\ \hline
\end{tabular}
}
\vspace{-0.0cm}
\end{table}

\begin{table}[t]
\caption{LUT count for a 512-input adder-tree.} \label{tab:resource}
\begin{tabular}{lccccc}
\hline
          & exact & \texttt{MAJ}   & \texttt{MAJ}-2 & over-feed & truncate \\ \hline
Synthesis & 638   & 183   & 116   & 383       & 340      \\
Equation  & 675   & 195   & 116   & 405       & 343      \\
Error     & 5.8\% & 6.6\% & 0.0\% & 5.7\%     & 0.9\%    \\ \hline
\end{tabular}
\vspace{-0.0cm}
\end{table}

\textbf{(2) Resource Utilization.}
To validate the efficiency of the proposed approximation techniques, in addition to holistic high-level performance and energy comparisons, we examine them by synthesizing a 512-input adder-tree.
Table \ref{tab:resource} represents the LUT utilization of the adder implemented in exact and approximate modes.
\texttt{MAJ}, \texttt{MAJ}-2, over-feed and truncate refer to the designs of Figure \ref{fig:Appr}(a)-(d).
It can be seen that our equations in Section \ref{subsec:hardware} have a modest average error of 3.8\%. Especially, it over-estimates the LUT count of both exact and approximate adders, so the resource \textit{saving} estimations remain similar to our predicted values.
For instance, synthesis results indicate \texttt{MAJ} (\texttt{MAJ}-2) saves 71.3\% (81.8\%) LUTs, which is very close to the predicted 71.1\% (82.8\%).

\begin{table}[t]
\caption{Relative accuracies SHEAR$er$ approximate encodings.} \label{tab:accuracy}
\resizebox{0.48\textwidth}{!}{
\begin{tabular}{lc|ccccc}
\hline
           & exact  & \texttt{MAJ}    & \texttt{MAJ}-2  & over-feed & trunc-3 & trunc-4 \\ \hline
\texttt{speech}     & 93.2\% & $-$0.7\% & $-$2.3\% & $-$0.8\%    & $-$0.9\%       & $-$1.9\%       \\
\texttt{activity}   & 93.9\% & $-$0.8\% & $-$1.2\% & $-$1.3\%    & $-$1.1\%       & $-$1.0\%       \\
\texttt{face}       & 95.5\% & $-$1.8\% & $-$3.3\% & $-$1.7\%    & $-$1.6\%       & $-$1.9\%       \\
\texttt{digit}      & 89.1\% & $-$0.8\% & $-$0.3\% & $-$1.7\%    & 0.1\%        & $-$0.1\%       \\ \hline
average      &  & $-$1.0\% & $-$1.8\% & $-$1.4\%    & $-$0.9\%        & $-$1.2\%       \\ \hline
LUT saving & 0      & 71.1\% & 82.8\% & 40.0\%    & 37.5\%       & 43.8\%       \\ \hline
\end{tabular}
}
\vspace{-0.0cm}
\end{table}

\textbf{(3) Accuracy.}
Table \ref{tab:accuracy} summarizes the accuracies of the proposed encodings relative to the exact encoding.
LUT saving, which is dataset-independent, is represented again for the comparison purpose.
``trunc-3'' and ``trunc-4'' stand for truncated encoding (Figure \ref{fig:Appr}(d)) where, respectively, three and four intermediate stages are truncated.
Overall, \texttt{MAJ} encoding (one-stage local majority shown in Figure \ref{fig:Appr}(a)) achieves an acceptable accuracy with significant resource saving, though it is not always the highest-accurate one.
For instance, in the \texttt{face} detection benchmark, the over-feed and 3-stage truncated encodings offer slightly better accuracy.
More interestingly, in the \texttt{digit} recognition dataset, trunc-4 shows a negligible $-0.1\%$ accuracy drop while trunc-3 even improves the accuracy by 0.1\%.
This can stem from the fact that emulating the hardware approximation in SHEAR$er$'s software layer takes a long time for the \texttt{digit} dataset, so we limited the software to try five different learning rate ($\alpha$) and repeat the entire training for five times (with $\texttt{epochs} = 50$) so the result might be slightly skewed.
For the other datasets we conducted the training for 25 times each with 50 $\texttt{epochs}$ to average out the variance of results.

\begin{figure}[t]
  \centering
  \vspace{-0.0cm}
  \includegraphics[width=0.95\columnwidth]{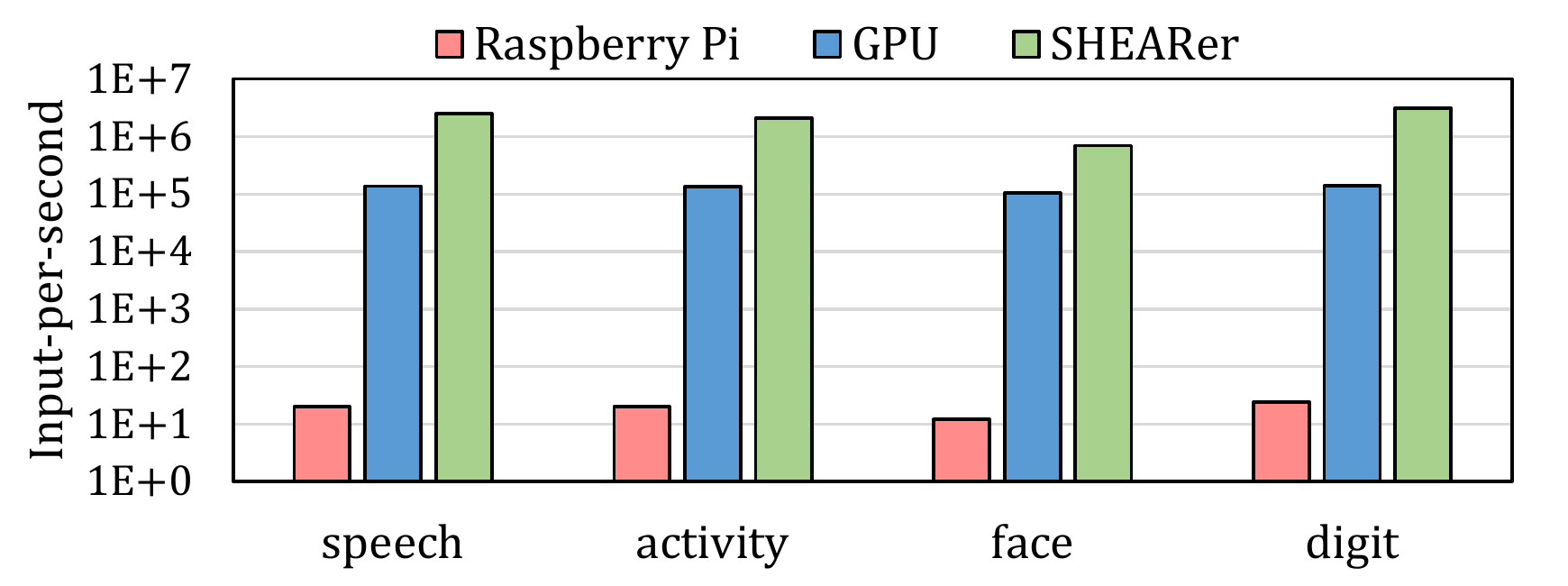} \vspace{-0.0cm}
  \caption{Throughput of SHEAR$er$ versus Raspberry Pi 3 and Nvidia GTX 1080 Ti. Y-axis is logarithmic scale.}\label{fig:throughput}  \vspace{-0.0cm}
\end{figure}

\textbf{(4) Performance.}
Figure \ref{fig:throughput} compares the throughput of SHEAR$er$ FPGA implementation with Raspberry Pi and Nvidia GPU.
SHEAR$er$ implementation is BRAM-bound, so all the exact and approximate implementations yield the same performance.
In Section \ref{subsec:arch} we elaborated that the \texttt{speech} dataset requires two cycles per $d_{mem}=64$ dimensions.
We can similarly show that \texttt{activity} and \texttt{digit} datasets also need two cycles per 64 dimensions, while \texttt{digit} requires three cycles as its level hypervectors are larger ($d_{hv} = \text{6,144}$) and occupy more BRAMs.
In the worst scenario, SHEAR$er$ improves the throughput by 58,333$\times$ and 6.7$\times$ compared to Raspberry Pi and GPU implementation. On average SHEAR$er$ provides a throughput of 104,904$\times$ and 15.7$\times$ as compared to Raspberry Pi and GPU, respectively.
The substantial improvements arise from that SHEAR$er$ adds up $\frac{d_{iv}}{2} \times 64$ (e.g., $\sim$25,000) numbers per cycle while also performs the binding (\texttt{XOR} operations) on the fly. However, Raspberry Pi executes sequentially and also its cache cannot fit all the class hypervectors with non-binary dimensions.
Note that we assume that dataset is available in the \textit{off-chip} memory (DRAM) of the FPGA. Otherwise, although per-sample latency would be affected, throughput remains the same as the off-chip memory latency is buried in the computation cycles.

\begin{figure}[t]
  \centering
  \includegraphics[width=1.0\columnwidth]{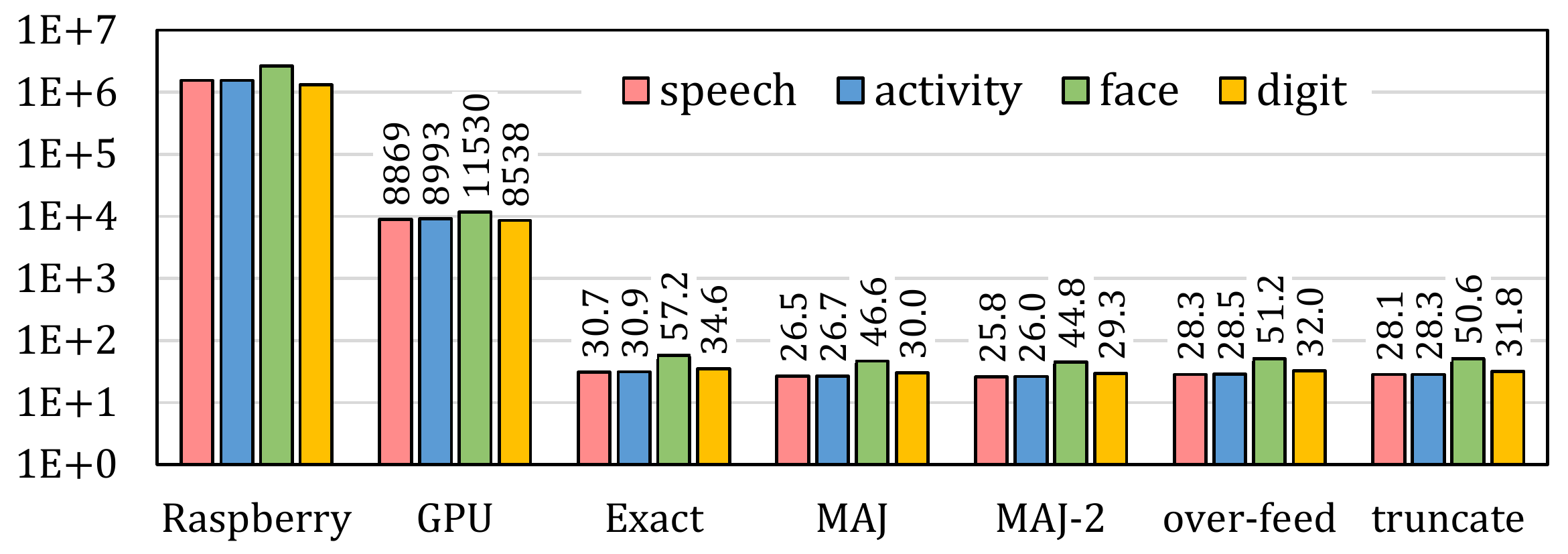} \vspace{-0.0cm}
  \caption{Energy (Joule) consumption of SHEAR$er$, Raspberry Pi and GPU for 10 million inference. Y-axis is logarithmic.}\label{fig:energy}  \vspace{-0.0cm}
\end{figure}

\textbf{(5) Energy Consumption.}
Figure \ref{fig:energy} compares the energy consumption of the exact and approximate SHEAR$er$ implementations with Raspberry Pi and GPU.
We have scaled the energy to 10 million inferences for the sake of illustration (Y-axis is logarithmic).
We used Hioki 3334 power meter and NVIDIA system management interface to measure the power consumption of Raspberry Pi and GPU, respectively.
We used Xilinx Power Estimator (XPE) to estimate the FPGA power consumption.
The average power of Raspberry Pi for all datasets hovers around 3.10 Watt, while this is $\sim$120 Watt for the GPU.
In FPGA implementation, powers showed more variation as the number of active LUTs and BRAMs differ between applications. E.g.,
The \texttt{face} dataset with two-stage majority encoding (\texttt{MAJ}-2) consumes 3.11 Watt, while the \texttt{digit} recognition dataset in the exact mode consumes 10.80 Watt. The smaller power consumption of \texttt{face} is mainly because of smaller off-chip data transfer as \texttt{face} has the largest hypervector length and takes 288 cycles to process an entire input, while for \texttt{digit} it takes 64 cycles.
On average, SHEAR$er$'s \textit{exact} encoding decreases the energy consumption of by 45,988$\times$ and 247$\times$ (average of all datasets) as compared to Raspberry Pi and GPU implementations.
\texttt{MAJ}-2 encoding of SHEAR$er$ consumes the minimum energy, which throttles the energy consumption by 56,044$\times$ and 301$\times$ compared to Raspberry Pi and GPU, respectively.
Note that power improvement of the approximate encodings is not proportional to their resource (LUT) utilization as BRAM power remains the same for all encodings.

\section{Conclusion} \label{sec:conc}
In this paper, we leveraged the intrinsic error resiliency of HD computing to develop different approximate encodings with varied accuracy and resource utilization attributes.
With a modest $1.0\%$ accuracy drop, our approximate encoding reduces the LUT utilization by 71.1\%.
By effectively utilizing the on-chip BRAMs of FPGA, we also proposed a highly efficient implementation that outperforms an optimized GPU implementation over 15$\times$, and surpasses Raspberry Pi by over five orders of magnitude.
Our FPGA implementation also consumes a moderate power: a minimum of 3.11 Watt for a face detection dataset using approximate encoding, and a maximum of 10.8 Watt on a digit recognition dataset when using exact encoding.
Eventually, our implementation reduces the energy consumption by 247$\times$ (45,988$\times$) compared to GPU and Raspberry Pi in exact encoding, which further improves by a factor of $1.22\times$ using approximate encoding.
\vspace{-0.0cm}

\section*{Acknowledgements}
This work was supported in part by CRISP, one of six centers in JUMP, an SRC program sponsored by DARPA, in part by SRC Global Research Collaboration (GRC) grant, DARPA HyDDENN grant, and NSF grants \#1911095 and \#2003279.

\bibliographystyle{ieeetr}
\bibliography{references}

\begin{thebibliography}{10}

\bibitem{kanerva2009hyperdimensional}
P.~Kanerva, ``Hyperdimensional computing: An introduction to computing in
  distributed representation with high-dimensional random vectors,'' {\em
  Cognitive computation}, vol.~1, no.~2, pp.~139--159, 2009.

\bibitem{masse2009olfactory}
N.~Y. Masse, G.~C. Turner, and G.~S. Jefferis, ``Olfactory information
  processing in drosophila,'' {\em Current Biology}, vol.~19, no.~16,
  pp.~R700--R713, 2009.

\bibitem{turner2008olfactory}
G.~C. Turner, M.~Bazhenov, and G.~Laurent, ``Olfactory representations by
  drosophila mushroom body neurons,'' {\em Journal of Neurophysiology},
  vol.~99, no.~2, pp.~734--746, 2008.

\bibitem{wilson2013early}
R.~I. Wilson, ``Early olfactory processing in drosophila: mechanisms and
  principles,'' {\em Annual Review of Neuroscience}, vol.~36, pp.~217--241,
  2013.

\bibitem{olshausen2004sparse}
B.~A. Olshausen and D.~J. Field, ``Sparse coding of sensory inputs,'' {\em
  Current Opinion in Neurobiology}, vol.~14, no.~4, pp.~481--487, 2004.

\bibitem{plate1995holographic}
T.~A. Plate, ``Holographic reduced representations,'' {\em IEEE Transactions on
  Neural networks}, vol.~6, no.~3, pp.~623--641, 1995.

\bibitem{wan2012web}
M.~Wan, A.~J{\"o}nsson, C.~Wang, L.~Li, and Y.~Yang, ``Web user clustering and
  web prefetching using random indexing with weight functions,'' {\em Knowledge
  and information systems}, vol.~33, no.~1, pp.~89--115, 2012.

\bibitem{kim2020genie}
Y.~Kim, M.~Imani, N.~Moshiri, and T.~Rosing, ``Geniehd: Efficient dna pattern
  matching accelerator using hyperdimensional computing,'' in {\em 2020 Design,
  Automation \& Test in Europe Conference \& Exhibition (DATE - To Appear)},
  IEEE, 2020.

\bibitem{imani2018hdna}
M.~Imani, T.~Nassar, A.~Rahimi, and T.~Rosing, ``Hdna: Energy-efficient dna
  sequencing using hyperdimensional computing,'' in {\em 2018 IEEE EMBS
  International Conference on Biomedical \& Health Informatics (BHI)},
  pp.~271--274, IEEE, 2018.

\bibitem{rahimi2018efficient}
A.~Rahimi, P.~Kanerva, L.~Benini, and J.~M. Rabaey, ``Efficient biosignal
  processing using hyperdimensional computing: Network templates for combined
  learning and classification of exg signals,'' {\em Proceedings of the IEEE},
  vol.~107, no.~1, pp.~123--143, 2018.

\bibitem{fatemeh2020embc}
F.~Asgarinejad, A.~Thomas, and T.~Rosing, ``Detection of epileptic seizures
  from surface eeg using hyperdimensional computing,'' in {\em 2020 42nd Annual
  International Conference of the IEEE Engineering in Medicine and Biology
  Society (EMBC) \emph{(to appear)}}, 2020.

\bibitem{mitrokhin2019learning}
A.~Mitrokhin, P.~Sutor, C.~Ferm{\"u}ller, and Y.~Aloimonos, ``Learning
  sensorimotor control with neuromorphic sensors: Toward hyperdimensional
  active perception,'' {\em Science Robotics}, vol.~4, no.~30, p.~eaaw6736,
  2019.

\bibitem{neubert2019introduction}
P.~Neubert, S.~Schubert, and P.~Protzel, ``An introduction to hyperdimensional
  computing for robotics,'' {\em KI-K{\"u}nstliche Intelligenz}, vol.~33,
  no.~4, pp.~319--330, 2019.

\bibitem{imani2019framework}
M.~Imani, Y.~Kim, S.~Riazi, J.~Messerly, P.~Liu, F.~Koushanfar, and T.~Rosing,
  ``A framework for collaborative learning in secure high-dimensional space,''
  in {\em 2019 IEEE 12th International Conference on Cloud Computing (CLOUD)},
  pp.~435--446, IEEE, 2019.

\bibitem{behnam2020hd}
B.~Khaleghi, M.~Imani, and T.~Rosing, ``Prive-hd: Privacy-preserved
  hyperdimensional computing,'' in {\em Proceedings of the 57th Annual Design
  Automation Conference \emph{(to appear)}}, 2020.

\bibitem{imani2019bric}
M.~Imani, J.~Morris, J.~Messerly, H.~Shu, Y.~Deng, and T.~Rosing, ``Bric:
  Locality-based encoding for energy-efficient brain-inspired hyperdimensional
  computing,'' in {\em Proceedings of the 56th Annual Design Automation
  Conference 2019}, p.~52, ACM, 2019.

\bibitem{rahimi2016robust}
A.~Rahimi, P.~Kanerva, and J.~M. Rabaey, ``A robust and energy-efficient
  classifier using brain-inspired hyperdimensional computing,'' in {\em
  Proceedings of the 2016 International Symposium on Low Power Electronics and
  Design}, pp.~64--69, 2016.

\bibitem{imani2017voicehd}
M.~Imani, D.~Kong, A.~Rahimi, and T.~Rosing, ``Voicehd: Hyperdimensional
  computing for efficient speech recognition,'' in {\em 2017 IEEE International
  Conference on Rebooting Computing (ICRC)}, pp.~1--8, IEEE, 2017.

\bibitem{imani2019sparsehd}
M.~Imani, S.~Salamat, B.~Khaleghi, M.~Samragh, F.~Koushanfar, and T.~Rosing,
  ``Sparsehd: Algorithm-hardware co-optimization for efficient high-dimensional
  computing,'' in {\em 2019 IEEE 27th Annual International Symposium on
  Field-Programmable Custom Computing Machines (FCCM)}, pp.~190--198, IEEE,
  2019.

\bibitem{salamat2019f5}
S.~Salamat, M.~Imani, B.~Khaleghi, and T.~Rosing, ``F5-hd: Fast flexible
  fpga-based framework for refreshing hyperdimensional computing,'' in {\em
  Proceedings of the ACM/SIGDA International Symposium on Field-Programmable
  Gate Arrays}, pp.~53--62, 2019.

\bibitem{imani2019quanthd}
M.~Imani, S.~Bosch, S.~Datta, S.~Ramakrishna, S.~Salamat, J.~M. Rabaey, and
  T.~Rosing, ``Quanthd: A quantization framework for hyperdimensional
  computing,'' {\em IEEE Transactions on Computer-Aided Design of Integrated
  Circuits and Systems}, 2019.

\bibitem{schmuck2019hardware}
M.~Schmuck, L.~Benini, and A.~Rahimi, ``Hardware optimizations of dense binary
  hyperdimensional computing: Rematerialization of hypervectors, binarized
  bundling, and combinational associative memory,'' {\em ACM Journal on
  Emerging Technologies in Computing Systems (JETC)}, vol.~15, no.~4,
  pp.~1--25, 2019.

\bibitem{imani2019binary}
M.~Imani, J.~Messerly, F.~Wu, W.~Pi, and T.~Rosing, ``A binary learning
  framework for hyperdimensional computing,'' in {\em 2019 Design, Automation
  \& Test in Europe Conference \& Exhibition (DATE)}, pp.~126--131, IEEE, 2019.

\bibitem{datta2019programmable}
S.~Datta, R.~A. Antonio, A.~R. Ison, and J.~M. Rabaey, ``A programmable
  hyper-dimensional processor architecture for human-centric iot,'' {\em IEEE
  Journal on Emerging and Selected Topics in Circuits and Systems}, vol.~9,
  no.~3, pp.~439--452, 2019.

\bibitem{isolet}
``Uci machine learning repository.''
  \url{http://archive.ics.uci.edu/ml/datasets/ISOLET}.

\bibitem{xilinx7}
``7 series fpgas data sheet.'' Data Sheet, February 2108.

\bibitem{imani2017exploring}
M.~Imani, A.~Rahimi, D.~Kong, T.~Rosing, and J.~M. Rabaey, ``Exploring
  hyperdimensional associative memory,'' in {\em 2017 IEEE International
  Symposium on High Performance Computer Architecture (HPCA)}, pp.~445--456,
  IEEE, 2017.

\bibitem{imani2019fach}
M.~Imani, S.~Salamat, S.~Gupta, J.~Huang, and T.~Rosing, ``Fach: Fpga-based
  acceleration of hyperdimensional computing by reducing computational
  complexity,'' in {\em Proceedings of the 24th Asia and South Pacific Design
  Automation Conference}, pp.~493--498, 2019.

\bibitem{ucihar}
``Uci machine learning repository.''
  \url{https://archive.ics.uci.edu/ml/datasets/human+activity+recognition+using+smartphones}.

\bibitem{lecun1998mnist}
Y.~LeCun, C.~Cortes, and C.~J. Burges, ``The mnist database of handwritten
  digits, 1998,'' {\em URL http://yann. lecun. com/exdb/mnist}, vol.~10, p.~34,
  1998.

\bibitem{griffin2007caltech}
G.~Griffin, A.~Holub, and P.~Perona, ``Caltech-256 object category dataset,''
  2007.

\end{thebibliography}

\end{document}